\documentclass[9pt]{article}

\usepackage{geometry}
\usepackage{graphicx} % more modern
\usepackage{subfigure} 
\usepackage{natbib}

\usepackage{algorithm}
\usepackage{algorithmic}

\usepackage{epstopdf}
\usepackage{times,bm}
\usepackage{algorithm}
\usepackage{algorithmic}[1]
\usepackage{graphicx}
\usepackage{subfigure,amssymb,mathrsfs,amsthm,amsfonts,amsmath}
\usepackage{amsmath,graphicx,caption,color,epsfig,float,pbox,tabularx,wrapfig,mathrsfs,multirow,array}
\newcommand{\bfu}{\mathbf{u}}
\newcommand{\bfv}{\mathbf{v}}

\newcommand{\bfI}{\mathbf{I}}

\newcommand{\calT}{\mathcal{T}}
\newcommand{\calN}{\mathcal{N}}

\newcommand{\RB}{\mathbb{R}}
\newcommand{\EB}{\mathbb{E}}

\newcommand{\CS}{\mathcal{S}}

\newtheorem{lemma}{Lemma}%[section]
\newtheorem{theorem}{Theorem}%[section]
\newtheorem{assumption}{Assumption}%[section]
\newtheorem{definition}{Definition}%[section]
\title{Approximate Inference via Clustering}

\author{Qianqian Song} % LEAVE BLANK FOR ORIGINAL SUBMISSION.

\begin{document}

\maketitle

\begin{abstract}
In recent years, large-scale Bayesian learning draws a great deal of attention. 
However, in big-data era, the amount of data we face is growing much faster than our ability to deal with it. 
Fortunately, it is observed that large-scale datasets usually own rich internal structure and is somewhat redundant. 
In this paper, we attempt to simplify the Bayesian posterior via exploiting this  structure. 
Specifically, we restrict our interest to the so-called ``well-clustered'' datasets and construct an \emph{approximate posterior} according to the clustering information. 
Fortunately, the clustering structure can be efficiently obtained via a particular clustering algorithm. 
When constructing the approximate posterior, the data points in the same cluster are all replaced by the centroid of the cluster. 
As a result, the posterior can be significantly simplified.
Theoretically, we show that under certain conditions the approximate posterior we construct is close (measured by KL divergence) to the exact posterior.
Furthermore, thorough experiments are conducted to validate the fact that the constructed posterior is a good approximation to the true posterior and much easier to sample from. 
\end{abstract}

\section{Introduction}

Bayesian learning are appealing in their ability to capture uncertainty in learned parameters and avoid overfitting.
In today's big data era, large-scale Bayesian inference has received a lot of attention and is widely applied in deep learning, graph learning, general machine learning and many applications, e.g., outlier detection~\cite{chen2015bridging,li2015preconditioned,hoffman2010online,xu2014distributed,fu2021mimosa,fu2021probabilistic,wei2020bitcoin,zhou2021network,fu2019pearl,li2020copod,zhao2021automatic}, where the number of data samples is usually very large.
What's worse, the size of data we face is growing much faster than our ability to deal with it.
At the same time, a great number of Bayesian sampling techniques were raised.
Among these methods, Markov Chain Monte Carlo (MCMC) methods are popular tools for performing exact inference via posterior sampling~\cite{chen2015bridging,fu2019pearl}. 
One major benefit of MCMC techniques is that they guarantee asymptotically exact recovery of the posterior distribution as the number of posterior samples grows.
However, they take a prohibitively long time when dealing with large-scale datasets, since for posterior with $N$ ($N$ is usually a large number) data points, they must perform $O(N)$ operations to draw a sample.

Motivated by highly complex models where MCMC algorithms and other Monte Carlo methods were too inefficient by far, approximate Bayesian sampling have emerged,  where the output cannot be considered as simulations from the genuine posterior, even under idealized situations of infinite computing power.
The mainstream approximate inference approaches include variational Bayes~\citep{jordan1999introduction,wainwright2008graphical} and Expectation Propagation (EP)~\citep{minka2001expectation}.
Both variational and EP methods aim to use a tractable form to approximate the exact posterior distribution. 
However, these approximate approaches own some limits.
Taking variational Bayes as an instance, the core idea behind variational methods is that we use a distribution $q$ from a restricted family to approximate the exact posterior distribution $p(\theta \vert X)$. 
Concretely, we minimize the Kullback-Leibler (KL) divergence between  the variational distribution $q$ and posterior distribution $p(\theta \vert X)$, i.e., solving the following problem: 
\begin{equation*}
\underset{\lambda}{\arg\min }\ \text{KL}(q(\theta \vert \lambda) \Vert p(\theta \vert X) ), \\
\end{equation*}
where $q(\theta \vert\lambda)$ is characterized by the  variational parameter $\lambda$. 
However, since $q$ distribution owns tractable form, they are not able to capture the variation in the posterior all the time, especially when the true posterior is highly complex~\citep{green2015bayesian}.

None of the above-mentioned approaches, however, have considered the \emph{internal structure} of the dataset. 
%Existing work have observed the redundancy in large-scale dataset~\cite{allen2016exploiting}. 
\cite{braverman2011streaming,allen2016exploiting,fu2019ddl} demonstrated that in many datasets the data vectors exhibit clustering structure.
Moreover, \cite{allen2016exploiting} use this structure to accelerate optimization procedure.
Accordingly, in this paper, we consider to exploit this clustering information in Bayesian sampling scenario.
For ease of exposition, we first list the main contributions of this paper as follows. 
\begin{itemize}
\item We construct an \emph{approximate posterior} based on the clustering structure of the dataset. 
Particularly, in the approximate posterior, the data samples in the same cluster are all replaced by the centroid of the cluster.
As a result, the posterior can be significantly simplified.
	
\item Theoretical analysis is provided to show that the KL divergence between  the true posterior and the constructed approximate posterior can be bounded.
	
\item Empirically, we conduct a series of experiment to validate the fact that the constructed posterior is a good approximation to the true posterior and much easier to sample from. 
Worth to mention that the clustering procedure is performed only once for all and is efficient.
\end{itemize}

%The remainder of the paper is organized as follows.
%Sections~\ref{sec:notation} and \ref{sec:formulation} present the notation frequently used in this paper and problem formulation, respectively.
%Section~\ref{sec:method} describes our algorithm and theoretical analysis.
%The empirical results are shown in Section~\ref{sec:experiment}.
In the following, we start by presenting the notation frequently used in this paper. 
Then we describe our method and theoretical analysis, followed by empirical studies. 
Finally, we conclude our work in Section~\ref{sec:concl}. 
The proofs of theoretical results are given in the Appendix.

\section{Notation}

Suppose that we have $N$ independent observations $X = \{x_1,\ldots,x_N\}$, 
let $\theta\in \mathbb{R}^d$ be the parameter vector of interest and $p(\theta)$ be the prior distribution. 
Likelihood function is denoted $p(X \vert \theta)$.  
The posterior distribution $p(\theta \vert X)$ satisfies the Bayes rule as 
\begin{equation}
\label{eqn:exact}
\begin{aligned}
p(\theta \vert X) = &\ \ \frac{1}{Z}{p(\theta)p(X\vert\theta)} \\
\propto &\ \ p(\theta)p(X\vert\theta)= p(\theta) \prod\limits_{i=1}^N p(x_i \vert \theta),
\end{aligned}
\end{equation}
where $Z = p(X) = \int_{\theta} p(\theta)p(X\vert\theta)d\theta$, is called normalizing constant of the posterior distribution.
It is independent of the parameter $\theta$, thus is usually ignored.

We use $\Vert\cdot \Vert$ to represent the $l_2$ norm for a vector and $\Vert \bfv \Vert = \sqrt{\sum_{i=1}^{d}{v_i^2}}$ holds for any $\bfv$ satisfying $\bfv \in \RB^d$.
In the follows we briefly introduce some key definitions that are useful throughout
the paper. 

\begin{definition}
	If we have two separate probability distributions $p(x)$ and $q(x)$ over the same random variable $x$, we can measure how different these two distributions are using the Kullback-Leibler (KL) divergence:
	\begin{equation*}
	\text{KL} (p(x)\Vert q(x)) = \EB_{x\sim p}\bigg[\ln \frac{p(x)}{q(x)}\bigg]= \int p(x)\ln \frac{p(x)}{q(x)} dx.
	\end{equation*}
\end{definition}
The KL divergence of any two probability distributions $p(x)$ and $q(x)$ is greater or equal to 0.
The equality is obtained if and only if $p = q$ almost everywhere.
In general KL divergence is intractable.
But when both $p$ and $q$ are Gaussian distribution, it can be computed analytically, shown as follows.
\begin{lemma}
	\label{lemma:1}
	Assume $p_1$ and $p_2$ are multivariate normal distribution with mean $\mu_1, \mu_2\in \RB^d$ and covariance $\Sigma_1,\Sigma_2 \in \RB^{d\times d}$.
	%That is, $ p_1(x\vert\mu_1,\Sigma_1) = (2\pi)^{-\frac{d}{2}} \vert\Sigma_1 \vert^{-\frac{1}{2}} \exp\big( - \frac{1}{2} (x-\mu_1)^\top\Sigma_1^{-1}(x-\mu_1)\big)$.
	%\begin{equation*} \begin{aligned}
	%& p_1(x\vert\mu_1,\Sigma_1) \\ = & (2\pi)^{-\frac{d}{2}} \vert\Sigma_1 \vert^{-\frac{1}{2}}  \exp\big( - \frac{1}{2} (x-\mu_1)^\top\Sigma_1^{-1}(x-\mu_1)\big).
	%\end{aligned} \end{equation*}
	Then we have 
	\begin{equation*}
	\begin{aligned}
	& \text{KL}(p_1(x\vert\mu_1,\Sigma_1) \Vert p_2(x\vert\mu_2,\Sigma_2))\\
	=&  - \frac{1}{2} \ln \vert \Sigma_1 \vert + \frac{1}{2} \ln \vert \Sigma_2 \vert  - \frac{1}{2} d \\
	& + \frac{1}{2} \big[ \text{tr}[\Sigma_2^{-1}  \Sigma_1] + (\mu_1 - \mu_2)^\top\Sigma_2^{-1} (\mu_1 - \mu_2) \big].\\
	\end{aligned}
	\end{equation*}
\end{lemma}

\begin{definition}
	A function $\phi: \RB^{d_1}\xrightarrow{}\RB^{d_2}$ is $L$-Lipschitz, if for all $\bfu,\bfv\in \RB^{d_1}$, we have
	\begin{equation*}
	\Vert\phi(\bfu) - \phi(\bfv) \Vert \leq L \Vert\bfu - \bfv \Vert.
	\end{equation*}
\end{definition}

\begin{definition}
	A function $\phi: \RB^d\xrightarrow{}\RB$ is ($1/\gamma$)-smooth if it is differentiable and its gradient is $(1/\gamma)$-Lipschitz, or, equivalently for all $\bfu,\bfv\in \RB^d$, we have
	\begin{equation*}
	\phi(\bfu) \leq \phi(\bfv) + \nabla \phi(\bfv)^\top (\bfu - \bfv) + \frac{1}{2\gamma}\Vert \bfu - \bfv \Vert^2.
	\end{equation*}
\end{definition}

Then we define some notations about the clustering information of the data vectors. 
Assume that in the clustering procedure, the data vectors $\{x_1,\ldots,x_N\}$ are divided into $c$ disjoint sets.
Each set corresponds to a cluster. 
The $j$-th cluster is denoted $\CS_j$ and owns $n_j$ data points, i.e., $n_j = \vert\CS_j \vert$. 
The sum of all $n_j$ is $N$, that is, $\sum_{j = 1}^{c} n_j = N$.
Then the \emph{centroid} and \emph{radius} of the cluster $\CS_j$ are defined as $ \mu_j = \frac{1}{n_j} \sum_{i\in \CS_j}^{} x_i$ and 
$\delta_j \triangleq \max_{i\in \CS_j} \Vert x_i - \mu_j \Vert$, respectively.
Correspondingly, the \emph{global radius} is defined as the maximal radius of all clusters, that is, $\delta = \max_{j=1}^c \delta_j$, which is also called \emph{radius} for simplicity. 
Furthermore, when the radius is small enough\footnote{usually much smaller than the distance between centroid of different clusters} and the number of clusters is far less than the number of data points, we usually call the dataset \emph{well-clustered}.
In this paper, we restrict our attention to cut down the redundancy and performing approximate inference in these ``well-clustered'' dataset.

\section{Methodology}

%Ironically, the primary cause MCMC methods are slow is that they are designed to be unbiased.
%If we were to allow a small bias in the stationary distribution, it is possible to simulate more cheaply~\citep{welling2011bayesian,korattikara2013austerity,chen2015convergence}.

To illustrate why internal structure can be helpful and gain some intuition for our idea, we provide an extreme case: 
if we assume that all the data vectors are located at the same spot, i.e., $x_1 = x_2 = \cdots = x_N$, then easy to find that the posterior can be rewritten into $p(\theta \vert X) \propto p(\theta) p(x_1\vert\theta)^N$. 
It means that the dataset is extremely redundant and we only require one data point to represent the posterior and perform the exact inference, while the other data points can be thrown away.

Now we relax the assumption, if we assume that data vectors forms clusters and the \emph{radius} of a certain cluster is infinitely small, a natural idea is to replace all the data points in the current cluster with only one data point (e.g., a natural choice is the centroid of the cluster).
The constructed approximate posterior would be close to the exact posterior.
% as long as the radius of the cluster is small enough.

Fortunately, this clustering structure is common in large-scale datasets~\citep{allen2016exploiting}.
%Taking the famous Covtype dataset as an instance, \cite{allen2016exploiting} demonstrated that 581,012 data points in Covtype can be efficiently categorized into 1,445 clusters of radius 0.1. 
%We also conduct a series of empirical studies to illustrate it.
Thus, in this paper, we focus on the so-called ``well-clustered'' datasets and attempt to exploit this information.
%The data points in the same cluster are all replaced by the centroid of the cluster.
Suppose we have already obtained the clustering structure of the dataset, i.e., we have known that data vectors $\{x_1,\cdots,x_N\}$ can be divided into $c$ disjoint sets, denoted  $\CS_1,\cdots,\CS_c$.  
We expect that the centroid of a certain cluster could provide a rough estimation for all the data points in the cluster and construct the approximate posterior based on this clustering structure.
Particularly, in the approximate posterior $\tilde{p}(\theta\vert X)$, the data samples in the same cluster are all replaced by the centroid of the cluster, described as follows:
\begin{equation}
\label{eqn:approx_post}
\begin{aligned}
\tilde{p}(\theta\vert X)  \propto &\ \ p(\theta) \prod_{j=1}^{c} \big( p(\mu_j \vert\theta)\big)^{n_j}\\
= &\ \ \frac{1}{\tilde{Z}} p(\theta) \prod_{j=1}^{c} \big( p(\mu_j \vert \theta )\big)^{n_j}, 
\end{aligned}
\end{equation}
where $\mu_j$ is the centroid of the $j$-th cluster $\CS_j$, $n_j$ corresponds to the cardinality of  $\CS_j $ and satisfy that $\sum_{j = 1}^{c} n_j = N$.
$\tilde{Z} = \int p(\theta) \prod_{j=1}^{c} \big( p(\mu_j \vert \theta )\big)^{n_j}d\theta$ represents the normalization constant of the approximate posterior.

Now we can see that sampling from the approximate posterior $\tilde{p}(\theta\vert X)$ is equivalent to sampling from the posterior $p(\theta \vert \tilde{X})$,
where the new dataset $\tilde{X}$ owns $N$ data points $\{ \underbrace{\mu_1,\ldots,\mu_1}_{n_1} , \cdots,  \underbrace{\mu_c,\ldots,\mu_c}_{n_c}  \}$,  or equivalently $c$ pseudo-data $\mu_1,\ldots,\mu_c$ and corresponding multiplicities $n_1,\ldots,n_c$.
The data is compressed, which means that we only need to store $c$ (pseudo) data point in the memory, instead of $N$.
The reduction on data size is significant when dealing with large-scale dataset. 
To measure how much redundancy is cut down, we introduce a concept called \emph{compression ratio}, defined as
\begin{equation}
\begin{aligned}
\rho  \triangleq \frac{c}{N}.
\end{aligned}
\end{equation}
Obviously, $\rho$ is always less than 1.
The memory cost in the approximate posterior is approximately\footnote{We need to store $\mu_1,\ldots,\mu_c$ and $n_1,\ldots,n_c$.} $\rho$ times of that in exact inference. 
In the following, we will see that during the sampling procedure the per-iteration running cost is also $\rho$ times of that in exact inference.

\subsection{Computational Overhead}
Now we discuss the computational overhead of drawing a sample using MCMC methods from the exact posterior (governed by Equation~\eqref{eqn:exact}) and the constructed approximate posterior (governed by Equation~\eqref{eqn:approx_post}).
For conventional random-walk based MCMC method, 
the main computational bottleneck is due to the Metropolis-Hastings (MH) step.
Concretely, computing the acceptance probability 
$\alpha = \min\{1,\frac{p(\theta)\prod_{i=1}^{N}p(x_i\vert \theta)}{p(\theta')\prod_{i=1}^{N}p(x_i\vert \theta')}\} $ (where $\theta$ represents the current state and $\theta'$ represents the proposed state) requires $O(N)$ operations. 
For gradient based MCMC approaches like Langevin Monte Carlo~\cite{rossky1978brownian} or Hamiltonian Monte Carlo~\cite{neal2011mcmc}, we have to additionally compute the gradient of log-posterior, given as $\nabla_{\theta} \log p(\theta) + \sum_{i=1}^{N} \nabla_{\theta}\log p(x_i \vert \theta)$.
This operation also involves $O(N)$ computations.

Contrarily, when sampling from the approximate posterior described in Equation~\eqref{eqn:approx_post}, the acceptance probability is equal to  
$\alpha = \min\{1,\frac{p(\theta) \prod_{j=1}^{c} \big( p(\mu_j \vert \theta )\big)^{n_j}}{p(\theta') \prod_{j=1}^{c} \big( p(\mu_j \vert \theta' )\big)^{n_j}}\} $.
Thus we only need $O(c)$ computations, so is the gradient of log-posterior when using gradient based MCMC approaches.
To conclude, drawing a sample using standard MCMC samplers from the approximate posterior requires only $O(c)$ operations. 
Although we introduce a small bias in the stationary distribution, which would be described later, we can use the computational time we save to draw more samples and reduce the variance.

%Furthermore, the large memory cost would be reduced when the amount of available data is very large. 
%Therefore, we can expect that the posterior can be significant easier if the data vectors are clustered.

\subsection{A particular clustering algorithm}

\begin{algorithm}[tb]
	\caption{Raw clustering algorithm}
	\label{alg:cluster}
	\begin{algorithmic}[1]
\REQUIRE $N$ data points $x_1,\cdots, x_N$, hyperparameter $\delta$
		%		\ENSURE $\theta_{p+1}$, $\theta_{p+2},\ldots$, 
\STATE $\calT \xleftarrow{}\{\}$.
\FOR {$i = 1,2, \ldots,N$}
\STATE $x_k \xleftarrow{} \text{Find-Nearest-Neighbor}(x_i, \calT, 2 \delta) $.	
\IF {$x_k$ does not exist }
\STATE add $x_i$ into $\calT$ as a new cluster.
\ELSE 
\STATE add $x_i$ into the cluster of $x_k$, denoted $\CS_j$.
\IF {the radius of $\CS_j$ exceed $\delta$}
\STATE remove $x_i$ from $\CS_j$, add $x_i$ into $\calT$ as a new cluster.
\ENDIF
\ENDIF 
\ENDFOR
	\end{algorithmic}
\end{algorithm}

To detect the clustering structure in the dataset, we resort to the well-studied approximate nearest neighbor algorithms.
We need to ensure that the selected approximate nearest neighbour algorithm is able to find a close neighbor with high probability if such a neighbor exists.
Fortunately, many efficient approximate nearest neighbor algorithms satisfy this requirement, say, Locality-Sensitive Hashing (LSH)~\citep{datar2004locality,andoni2015practical}, product quantization~\citep{jegou2011product}. 
In this paper, following \cite{allen2016exploiting}, we use E2LSH~\footnote{Details about E2LSH can be found in http://www.mit.edu/~andoni/LSH/}.
Based on this, we devise a particular clustering algorithm, which is listed in Algorithm~\ref{alg:cluster}.

Concretely, given $N$ data points $x_1,\ldots,x_N$, we iteratively call $\text{Find-Nearest-Neighbor}$ for each $i = 1,2,\ldots,N$. 
Find-Nearest-Neighbor$(x_i,\calT,2\delta)$ is an oracle that either a close neighbor of $x_i$ with distance at most $2\delta$ in the set $\calT$ or nothing, where $\calT$ contains the clustering information of $\{x_1,\cdots,x_{i-1} \}$.
If Find-Nearest-Neighbor$(x_i,\calT,2\delta)$ returns nothing, which means $x_i$ does not belong to any existing cluster, then we would create a new cluster in $\calT$ for $x_i$ (Step 5 in Algorithm~\ref{alg:cluster}).
If Find-Nearest-Neighbor$(x_i,\calT,2\delta)$ returns a neighbor, we try to add $x_i$ into this cluster and recompute the radius of the new cluster.
We claim $x_i$ \emph{belongs} to this cluster if the radius does not exceed $\delta$. 
Otherwise, we remove $x_i$ from the cluster and add it into $\calT$ as a new cluster (Step 9). 
As a consequence, given the radius we want, the algorithm is able to output the clustering result of the dataset. 
Thus, we can obtain different clustering information via adjusting the knob (i.e., radius $\delta$).
The empirical effect of the various knobs would be empirically studied later in Section~\ref{sec:experiment}.

%\paragraph{tolerance} may fail to find a neighbor 
%For example,  But it can ensure that 

\subsection{Theoretical results}
\label{sec:theory}
Now we explore the theoretical properties of the constructed approximate posterior $\tilde{p}(\theta \vert X)$. 
Our goal is to show $\tilde{p}(\theta \vert X)$ is close to ${p}(\theta \vert X)$.
On the other hand, as stated before, KL divergence is a mainstream metric that measure the difference between two probability distributions.
Thus, we attempt to prove that for well-clustered dataset, the KL divergence between the true posterior and the approximate posterior can be bounded. 
First, we introduce some gentle assumptions. 

\begin{assumption}
\label{asmp:smooth}
($1/\gamma$-smoothness condition) The log-likelihood function $\ln p(\cdot \vert \theta): \RB^d \xrightarrow{} \RB $ is ($1/\gamma$)-smooth for any given  $\theta$. 
\end{assumption}

\begin{assumption}
\label{asmp:lipschitz}
(Lipschitz condition) There exists two constant $0 < L_1,L_2 < \infty$ such that the log-likelihood function $\ln p(\cdot \vert \theta): \RB^d \xrightarrow{} \RB $, is ($L_1$)-Lipschitz for any given $\theta$,
	the gradient of log-likelihood, i.e., $\nabla_{\theta} \ln p(\cdot \vert \theta): \RB^d \xrightarrow{} \RB^{d} $,  is ($L_2$)-Lipschitz for any given $\theta$.
\end{assumption}
These assumptions are widely used~\citep{johnson2013accelerating,zhao2014accelerating} and mild for many models, such as Bayesian logistic regression and Bayesian linear regression used in Section~\ref{sec:experiment}.
Then the main theoretical result of this paper is shown as follows.
\begin{theorem}
	\label{thm:main}
	Under Assumption~\ref{asmp:smooth}, the KL divergence between the true posterior distribution and the approximate posterior described in Equation~\eqref{eqn:approx_post} can be bounded as: 
\begin{equation}
\label{eqn:bound}
\begin{aligned}
\text{KL} ( p(\theta \vert X)\Vert \tilde{p}(\theta\vert X) ) \leq K_0 \delta^2,
\end{aligned}
\end{equation}
where $K_0 = 2L_1L_2N + \frac{N}{\gamma}$ is a constant, independent of the clustering information of the dataset.
\end{theorem}

From the theorem, we know that the approximate posterior can be arbitrarily close to the exact posterior as the radius $\delta\xrightarrow{} 0$, which is consistent with our intuition. 
In other words, when $\delta = 0$, $\tilde{p}(\theta\vert X)$ reduces to $p(\theta \vert X)$. 
% It is obvious that when $\delta=0$, the approximate posterior reduces to the exact posterior. 
%As shown later in Section~\ref{sec:experiment}, $\delta$ usually takes a small value, like $0.01$. 
%In this case, the approximate posterior is extremely close to the exact posterior. 

\section{Empirical Evaluation}
\label{sec:experiment}

In this section, we investigate the empirical performance of the proposed approach,
%samples drawn from approximate posterior $\tilde{p}(\theta \vert X)$.
which mainly consists of two steps and is easy-to-use.
Firstly, the clustering procedure is performed to the data vectors ahead of sampling procedure.
According to the clustering results, we construct the approximate posterior $\tilde{p}(\theta \vert X)$.
In the second step, we use sampler to draw samples from the approximate posterior $\tilde{p}(\theta \vert X)$.
Accordingly, the empirical studies can be divided into two parts.
In the first part, we investigate the scalability of the clustering method we use and show that the clustering procedure can be finished in a relatively short time, compared with the subsequent sampling procedure.
Then, in the second part, we report the performance of both exact posterior and approximate posterior using the same sampler.
In this paper, we select Hamiltonian Monte Carlo (HMC)~\citep{neal2011mcmc}, a state-of-the-art MCMC method, as the standard sampler. 
The step size $\epsilon$ and the number of leapfrog steps $L$ are two parameters of HMC.
In this paper, number of leaps is fixed at $10$ for all the datasets and we tune the stepsize for each tasks.
Moreover, the radius $\delta$ plays the role of knob parameter in the clustering procedure.
Thus, we would explore the empirical impact of various knobs. 
%We find that there is a tradeoff 

As stated above, we use the same sampler to draw from exact (baseline) and different approximate posteriors (corresponding to various knobs).
Thus it is senseless to compare the posterior value of samples drawn from different posteriors.
Instead, we choose to monitor the test error and record it as a function of running time to measure the efficiency of the methods.

Furthermore, in order to observe 
the difference between $\tilde{p}(\theta \vert X)$ and $p(\theta \vert X)$,
it is desirable to check the KL divergence $\text{KL}(p(\theta \vert X)\Vert \tilde{p}(\theta \vert X))$.
However, in the context of Bayesian inference, 
$\text{KL}(p(\theta \vert X)\Vert \tilde{p}(\theta \vert X))$ is intractable analytically in the general case. 
To handle this challenge, we follow \cite{xu2014distributed,li2015stochastic} measuring the performance by computing an empirical estimation about $\text{KL}(p(\theta \vert X)\Vert \tilde{p}(\theta \vert X))$, where both $p(\theta \vert X)$ and $\tilde{p}(\theta \vert X)$ are replaced by a Gaussian
\footnote{If both $p$ and $q$ are Gaussian, $\text{KL}(p\Vert q)$ is tractable analytically, as shown in Lemma~\ref{lemma:1}.} 
that had the same mean and covariance as samples drawn from the posterior using certain MCMC sampler (HMC here), to quantify the calibration of uncertainty estimations.
Moreover, as mentioned before, variational Bayes~\citep{jordan1999introduction,wainwright2008graphical} is a mainstream approximate inference technique. 
Thus, to show how well $\tilde{p}(\theta \vert X)$ approximates $p(\theta \vert X)$, 
%to show the effectiveness of our approximate posterior $\tilde{p}(\theta \vert X)$, 
we also list the empirical estimation to the KL divergence between exact posterior ${p}(\theta \vert X)$ and variational posterior $q(\theta\vert \lambda)$ as a reference.
We focus on applying our approach to popular machine learning tasks, carried out on benchmark datasets: (i) Bayesian logistic regression (ii) Bayesian linear regression.
All of the datasets used in Bayesian logistic regression and Bayesian linear regression model can be downloaded from the LIBSVM website\footnote{http://www.csie.ntu.edu.tw/~cjlin/libsvmtools/}.
Details about these datasets 
%and some settings 
can be found in Table~\ref{table:logistic_dataset} and \ref{table:linear_regression}.
They are chosen 
%fairly randomly in order
to cover various sizes of datasets.
To normalize the data, we define $R = \max_{i=1}^{N} \Vert x_i\Vert_2$, and all the feature vectors are divided by $R$.
Such an operation would guarantee that $\Vert x_i\Vert_2 \leq 1 $ for $\forall\ i = 1,\cdots,N$, that is, all the feature vectors lie in the unit sphere.
Accordingly, the radius of cluster would not be too large.
%\subsection{Toy model}
% SVRG-LD

\subsection{Bayesian logistic regression}
\label{sec:logistic}
Here, we restrict our interest on Bayesian multiclass logistic regression. 
Let $x\in\RB^d$ be a vector of feature values and $y=[y_1,\ldots,y_K]^{\top} \in\RB^K$ be a $K$-dimensional $0/1$ valued vector, where $K$ is the number of classes. There exists only one $k\in\{1,\ldots,K\}$ such that $y_k=1$ while other coordinates are all 0.
Multiclass logistic regression is a conditional probability model of the form 
\begin{equation*}
\begin{aligned}
p(y_k = 1\vert x,W) = \frac{\exp(w_k^{\top} x)}{\sum\nolimits_{j=1}^K\exp(w_{j}^{\top} x)}
\end{aligned}
\end{equation*}
parametrized by the matrix $W = [w_1,w_2,\ldots,w_K]\in\RB^{d\times K}$.
Each column of $W$ corresponds to one class.
The Gaussian prior is used, encouraging all the elements of  $W$ near 0. 
In variational methods, we proposed a variational Gaussian distribution $q(W\vert \bf{\mu},\Sigma)$ to approximate the intractable posterior of regression parameter. 
Further, we assume that the covariance matrix $\Sigma$ is diagonal for efficiency and feasibility. 

\subsection{Bayesian linear regression}

In Bayesian linear regression, we are provided with $N$ data samples, $\{x_i, y_i\}_{i=1}^{N}$, where $x_i \in \RB^d$ is a $d$-dimensional feature vector and $y_i \in \RB$ represents the target value.
The distribution of the $i$-th output $y_i$ is given by $p(y_i \vert x_i) = \calN (\beta^\top x_i, \gamma)$, where $\beta \in \RB^d$ is the parameter of interest.
Gaussian prior is employed, i.e., the prior distribution satisfy that $p(\beta) = \calN(0,\lambda \bfI)$.
Owing to the conjugacy, the posterior distribution over $\beta$ is also a Gaussian distribution.
The gradient of log-likelihood is $\nabla_{\beta} \log p(y_i \vert x_i) = - (y_i - \beta^\top x_i) x_i $.
For variational Bayes methods, similar to Section~\ref{sec:logistic}, we use a Gaussian with diagonal covariance to approximate the true posterior.

\begin{table}[]
\centering\small 
\begin{tabular}{|c|c|c|c|c|c|}
\hline
dataset & class & training/testing size & feature dim  \\ \hline
SensIT (acoustic) & 3 & 78,823 / 19,705 & 50 \\ \hline 
SensIT (seismic) & 3 & 78,823 / 19,705 & 50 \\ \hline 
covtype.binary & 2 & 523,124/57,888 & 54   \\ \hline		
mnist & 10 & 60,000/10,000 & 784  \\ \hline	
%news20 & 20 & 15,935 / 3,993 & 62,061 \\ \hline 
%poker & 10 & 25,010 / 1,000,000 & 10 \\ \hline 
%sensorless & 11 & 58,509 & 48 \\ \hline 
%shuttle & 7 & 43,500 / 14,500 & 9 \\ \hline 
%cifar & 10 & 50,000/10,000 & 1024 \\ \hline	
%usps & 10 & 7,291/2,007 & 256  \\  \hline
%pendigits & 10 & 7,494/3,498  & 16   \\	\hline
\end{tabular}
\caption{Bayesian Logistic Regression: the datasets}
\label{table:logistic_dataset}
\end{table} 

\subsection{Results and Analysis}

Now we show the empirical results. 
% Firstly, for various approximate posteriors (including the exact posterior as baseline), we report the test error as a function of running time in Figure~\ref{fig:logistic}. 
% For each dataset, we choose three representative settings to report.
% From the results, we see that the test error can converge to a low value more rapidly when sampling from the approximate posterior.
% Furthermore, 
we report results about the approximate posterior in Table~\ref{table:all}, especially the clustering information, including (i) number of clusters $c$; (ii) radius $\delta$; (iii) compression ratio $\rho$.
Due to the randomness of the algorithms, the reported results are the average of 5 independent trials.
We observe that given a reasonable radius $\delta$, the approximate KL divergence between the exact posterior $p(\theta \vert X)$ and the approximate posterior $\tilde{p}(\theta \vert X)$ is relatively small, compared with $\text{KL}( p(\theta \vert X) \Vert q(\theta\vert\lambda) )$, i.e., the KL divergence between the true posterior distribution and the variational distribution. 
Thus, we claim that $\tilde{p}(\theta \vert X)$ is a better approximation to the true posterior than variational distribution in this case.

Furthermore, we find that our method achieves significant acceleration as well as a data size reduction (less memory cost).
Large-scale dataset can always be greatly compressed without significant performance degradation.
%For example, for Covtype dataset, 
It is worth noting that the dataset compression relies heavily on the redundancy of the dataset.
%But it is reasonable to claim that it is applicable to most of the selected dataset.
%Usually a compression ratio at $30-50\%$ will not 

We also find that the clustering procedure costs much less running time than the subsequent sampling procedure. 
The clustering procedure serves as an efficient preprocessing step to the dataset and reduce the redundancy in the original large-scale dataset.

From Table~\ref{table:all}, we also observe that for almost all the dataset, the radius $\delta$ is within the same order of magnitude, which allows us to select a small finite set to explore. 
Therefore it is easy to find an appropriate $\delta$.
We attribute it to the normalization step. 

\begin{figure}[]
	\subfigure[SensIT(acoustic), $\delta= 0.2$]{
		\centering
		\includegraphics[width=3.7cm]{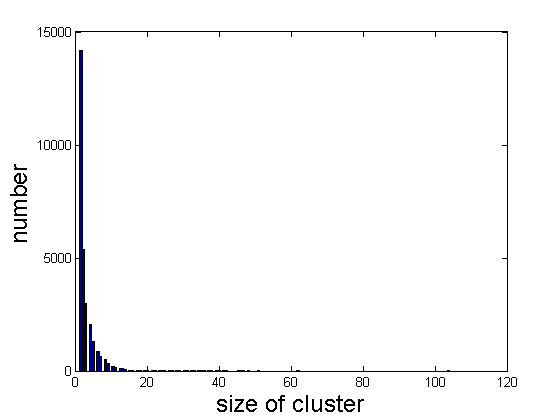}
		\label{fig:aaa}}
	\subfigure[mnist, $\delta = 0.1$]{
		\centering
		\includegraphics[width=3.7cm]{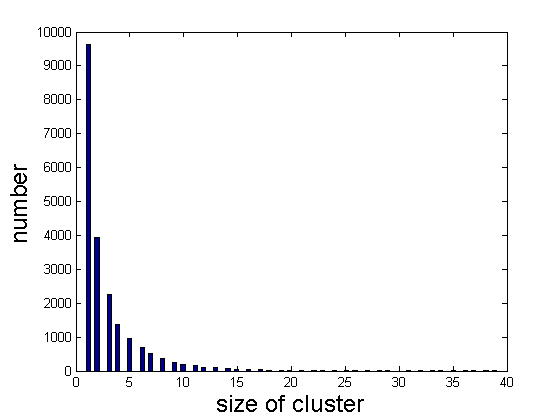}
		\label{fig:bbb}}
	\caption{Bayesian logistic regression: the distribution of cluster size.}	
	\label{fig:power}
\end{figure}

What's more, we find the size of cluster usually follows power-law distribution.
As an illustration, we randomly choose two datasets and plot the histogram about the size of cluster in Figure~\ref{fig:power}.

% \begin{figure*}[]
% 	\subfigure[SensIT(acoustic)]{
% 		\centering
% 		\includegraphics[width=4.7cm]{acoustic.eps}
% 		\label{fig:a}}
% 	\subfigure[SensIT(seismic)]{
% 		\centering
% 		\includegraphics[width=4.7cm]{seismic.eps}
% 		\label{fig:b}}
% 	\subfigure[covtype]{
% 		\centering
% 		\includegraphics[width=4.7cm]{covtype.eps}
% 		\label{fig:c}}
% %	\subfigure[mnist]{
% %		\centering
% %		\includegraphics[width=4.99cm]{mnist.eps}
% %		\label{fig:d}}		
% 	\caption{Bayesian logistic regression: test error vs running time.}	
% 	\label{fig:logistic}
% \end{figure*}

\begin{table}[]
\centering\small 
\begin{tabular}{|c|c|c|c|c|}
\hline
dataset & training/testing size & feature dim  \\ \hline
		%cifar & 50,000/10,000 & 1024 \\ \hline	
cadata & 16,512/4,628 & 8 \\ \hline 
%cpusmall & 6,554/1,638 & 12 \\ \hline 
YearPredictionMSD & 463,715 / 51,630 & 90 \\ \hline 
E2006 & 16,087 / 3,308 & 150,360 \\ \hline 
%		E2006-log1p & 16,087 / 3,308 &  4,272,227 \\ \hline 
\end{tabular}
\caption{Bayesian linear Regression: the datasets}
\label{table:linear_regression}
\end{table}

\section{Conclusion}
\label{sec:concl}

In this paper, we attempt to exploit the internal structure of large-scale dataset in the context of Bayesian sampling.
Particularly, we have devised an easy-to-use scheme for so-called ``well-clustered'' dataset, that is, firstly an efficient clustering procedure is implemented to obtain the clustering information of the data points, then we construct an approximate posterior based on the clustering information and draw samples from the approximate posterior. 
Compared with the exact posterior, the approximate posterior is significantly simplified and much easier to sample from. 
Theoretical analysis has been provided to guarantee that the KL divergence between the exact and approximate posterior can be bounded.
Furthermore, the empirical evaluations are exhaustive to backup both the effectiveness and efficiency of such an easy-to-use scheme.

\begin{table*}[]
\centering
\caption{Results for both Bayesian logistic regression and Bayesian linear regression. 
Data size $N$,  number of clusters $c$,  radius $\delta$, compression ratio $\rho$, approxKL, test error and running time for each setting are listed.
approxKL is short for approximate KL divergence.
For $\tilde{p}(\theta \vert x)$, approxKL is an estimation to $\text{KL}(p(\theta \vert X)\Vert \tilde{p}(\theta \vert X))$ while for variational methods, approxKL is an estimation to $\text{KL}(p(\theta \vert X)\Vert q(\theta \vert \lambda))$.
Running time is measured in terms of seconds. 
For our method, running time is represented as $t_1 + t_2$, where $t_1$, $t_2$ represent the running time of clustering procedure and sampling procedure, respectively. 
Test error corresponds the classifying error rate on test set (\%) for logistic regression and test MSE for linear regression. 
For E2006, feature dimention $d$ is too large, so approxKL is hard to estimate. 
}
\label{table:all}
	\begin{tabular}{|l|c|c|c|c|c|c|c|}
	\hline
dataset & data num $N$ & clusters num $c$ & $\delta$  &  $\rho$ & approxKL  & test error & running time \\ \hline

\multirow{5}{*}{SensIT(acoustic)} & \multirow{5}{*}{78,823} & \multicolumn{4}{c|}{baseline method: exact inference}  & 32.02\% & 45.24 \\ \cline{3-8}
&  & 19,795 & 0.14 & 25.11\% & 1.92e04 & 32.02\% & 1.03 + 23.92 \\ \cline{3-8}
& & 10,543 & 0.20 & 13.38\% & 2.16e04 & 32.25\% & 0.92 + 17.43 \\ \cline{3-8}
& & 7,497 & 0.25 & 9.51\% & 2.87e04 & 32.27\% & 0.65 + 14.33\\ \cline{3-8}
& & \multicolumn{3}{c|}{variational Bayes}  & 8.74e04 & 32.50\% & 12.43 \\ \hline\hline

\multirow{5}{*}{SensIT(seismic)} & \multirow{5}{*}{78,823} &
\multicolumn{4}{c|}{baseline method: exact inference}  & 30.74\% & 36.63 \\ \cline{3-8}
& & 62,507 & 0.20 & 79.30\% & 2.52e04 & 30.73\% & 0.97+29.44 \\ \cline{3-8}
& & 23,537 & 0.25 & 29.86\% & 2.73e04 & 31.08\% & 1.00+18.23 \\ \cline{3-8}
& & 11,224 & 0.29 & 14.24\% & 3.23e04 & 31.30\% & 0.72+12.42 \\ \cline{3-8}
& & \multicolumn{3}{c|}{variational Bayes}  & 4.53e04 & 31.96\% & 10.42 \\ \hline \hline
		
\multirow{5}{*}{covtype} & \multirow{5}{*}{523,124} & \multicolumn{4}{c|}{baseline method: exact inference}  & 34.35\% & 143.82 \\ \cline{3-8}
& & 283,690 & 0.05 & 54.23\% & 0.92e04 & 34.64\% & 2.54+86.80 \\ \cline{3-8}
& & 150,668 & 0.10 & 28.80\% & 1.01e04 & 35.01\% & 1.86+45.52 \\ \cline{3-8}
& & 62,409  & 0.20 & 11.93\% & 1.62e04 & 34.89\% & 1.05+47.34 \\ \cline{3-8}
& & \multicolumn{3}{c|}{variational Bayes}  & 2.75e04 & 35.45\% & 24.48 \\ \hline \hline
		
\multirow{5}{*}{mnist} & \multirow{5}{*}{60,000} & \multicolumn{4}{c|}{baseline method: exact inference} 
& 14.58\% & 65.48 \\ \cline{3-8}
& & 22,890 & 0.08 & 38.15\% & 1.64e06 & 14.90\% & 1.95+45.43 \\ \cline{3-8}
& & 14,286 & 0.10 & 23.81\% & 2.31e06 & 14.98\% & 1.42+26.44 \\ \cline{3-8}
& & 9,458 & 0.15 & 15.76\% & 3.55e06 & 15.89\% & 1.22+21.84 \\ \cline{3-8}
& & \multicolumn{3}{c|}{variational Bayes}  & 6.62e06 & 19.85\% & 19.64 \\ \hline \hline
		
% \multirow{5}{*}{cadata} & \multirow{5}{*}{16,512} & \multicolumn{4}{c|}{baseline method: exact inference}  & 7.38e04 & 156.4  \\ \cline{3-8}
% & & 11,984 & 0.25 & 72.58\% & 1.83e02 & 7.45e04 & 10.89+90.43 \\ \cline{3-8}
% & & 7,559 & 0.3 & 45.78\% & 2.43e02 & 7.47e04 & 7.57+74.41 \\ \cline{3-8}
% & & 5,918 & 0.35 & 35.84\% & 2.73e02 & 7.48e04 & 6.43+66.32 \\ \cline{3-8}
% & & \multicolumn{3}{c|}{variational Bayes}  & 1.03e03 & 7.63e04 & 32.83 \\ \hline \hline
% % one pass 0.009 epsilon = 10^{-5}  4K burn-in 
		
% \multirow{5}{*}{YearPredictionMSD} & \multirow{5}{*}{463,715} & \multicolumn{4}{c|}{baseline method: exact inference}  & 51.85 & 137.2 \\ \cline{3-8}
% & & 170,794 & 0.5 & 36.83\% & 5.83e04 & 52.89 & 4.31+49.54 \\ \cline{3-8}
% & & 84,267 & 0.6 & 18.17\% & 6.18e04 & 53.10 & 3.20+35.84 \\ \cline{3-8}
% & & 51,243 & 0.7 & 11.05\% & 9.04e04 & 54.07 & 2.81+29.65 \\ \cline{3-8}
% & & \multicolumn{3}{c|}{variational Bayes}  & 1.62e05 & 53.09 & 36.45 \\ \hline \hline
% % one pass 0.38s    	
		
% \multirow{5}{*}{E2006} & \multirow{5}{*}{16,087} & \multicolumn{4}{c|}{baseline method: exact inference}  & 0.4913 & 2180.0 \\ \cline{3-8}
% & & 13,084 & 0.3 & 81.33\% & - & 0.4932 & 24.85+1927.5 \\ \cline{3-8}
% & & 7,807 & 0.6 & 48.53\% & - & 0.4947 & 19.01+1414.5 \\ \cline{3-8}
% & & 6,248 & 0.7 & 38.84\% & - & 0.4934 & 17.89+1062.4 \\ \cline{3-8}
% & & \multicolumn{3}{c|}{variational Bayes}  & - & 0.4957 & 841.2 \\ \hline 
% exact one pass time average: 5.67   150 iter burn-in		epsilon = 10^{-4} L = 10
	\end{tabular}
\end{table*}

\section*{Appendix}

\section*{Proof of Theorem~\ref{thm:main}}

\begin{proof}
	
First, we expand the KL divergence between the exact posterior and the approximate posterior as:
\begin{equation}
\begin{aligned}
\text{KL} ( p(\theta \vert X)\Vert \tilde{p}(\theta\vert X) ) = \int p(\theta \vert X) \ln \frac{p(\theta \vert X)  }{\tilde{p}(\theta\vert X)} d\theta. 
\end{aligned}
\end{equation}
	
Then we focus the term $\ln \frac{p(\theta \vert X)  }{\tilde{p}(\theta\vert X)}$ in the above equation and expand it as: 
\begin{equation}
\label{eqn:log_term}
\begin{aligned}
& \ln \frac{p(\theta \vert X)  }{\tilde{p}(\theta\vert X)}\\
= & \ln \frac{ \frac{1}{Z} p(\theta) \prod_{i=1}^{N} p(x_i \vert\theta) }{\frac{1}{\tilde{Z}} p(\theta) \prod_{j=1}^{c} \big( p(\mu_j \vert \theta )\big)^{n_j}}\\
= & \ln \frac{ p(\theta) \prod_{i=1}^{N} p(x_i \vert\theta) }{ p(\theta) \prod_{j=1}^{c} \big( p(\mu_j \vert \theta )\big)^{n_j}} + \ln \frac{\tilde{Z}}{Z} \\
= &  \sum_{j=1}^{c} \ln \frac{\prod_{ i \in \CS_j}^{} p(x_i \vert \theta)}{\big( p(\mu_j \vert \theta)\big)^{n_j}} + \ln \frac{\tilde{Z}}{Z}\\
= & \bigg[ \sum_{j=1}^{c}  \sum_{i \in \CS_j}^{}  \big[\ln p(x_i \vert \theta)\big] -  n_j \ln p(\mu_j \vert \theta)\bigg] + \ln \frac{\tilde{Z}}{Z} \\
= & \sum_{j=1}^{c} \underbrace{\bigg[ \sum_{i \in \CS_j}^{}  \big[\ln p(x_i \vert \theta) -  \ln p(\mu_j \vert \theta) \big] \bigg]}_{A} + \underbrace{\ln \frac{\tilde{Z}}{Z}}_{B}. \\ 
\end{aligned}
\end{equation}

Then we consider the term ``A'' in the above equation: $\sum_{i \in \CS_j}^{}  \big[\ln p(x_i \vert \theta) -  \ln p(\mu_j \vert \theta) \big]$ and try to bound it, which can be divided into two parts.
The first part is to upperbound $\sum_{i \in \CS_j}^{}  \big[\ln p(x_i \vert \theta) -  \ln p(\mu_j \vert \theta) \big]$ while the second part is to upperbound $\sum_{i \in \CS_j}^{}  \big[  \ln p(\mu_j \vert \theta) - \ln p(x_i \vert \theta) \big]$:	
	
(i) Part I:	
\begin{equation}
\label{eqn:term_a}
\begin{aligned}
&  \sum_{i \in \CS_j}^{}  \big[\ln p(x_i \vert \theta) -  \ln p(\mu_j \vert \theta) \big] \\
\leq &  \sum_{i \in \CS_j}^{}  \big[\ln p(\mu_j \vert \theta) + \nabla \ln p(\mu_j \vert \theta)^\top (x_i - \mu_j )\\
&  + \frac{1}{2\gamma}\Vert x_i - \mu_j\Vert^2-  \ln p(\mu_j \vert \theta) \big]  \\ 
= &  \sum_{i \in \CS_j}^{} \frac{1}{2\gamma}\Vert x_i - \mu_j\Vert^2\\ 
\leq & \frac{n_j}{2\gamma}\max_{i \in \CS_j}\Vert x_i - \mu_j \Vert^2,
\end{aligned}
\end{equation}
where the first inequality follows from Assumption~\ref{asmp:smooth} and the first equality follows from the fact that $\frac{1}{n_j}\sum_{i\in\CS_j}^{}x_i = \mu_j$.

(ii) Part II:		
\begin{equation}
\label{eqn:term_a2}
\begin{aligned}
&  \sum_{i \in \CS_j}^{}  \big[  \ln p(\mu_j \vert \theta) - \ln p(x_i \vert \theta) \big] \\
\leq &  \sum_{i \in \CS_j}^{}  \big[\ln p(x_i \vert \theta) + \nabla \ln p(x_i \vert \theta)^\top ( \mu_j - x_i)\\
&  + \frac{1}{2\gamma}\Vert \mu_j - x_i\Vert^2-  \ln p(x_i \vert \theta) \big]  \\ 
= &  \sum_{i \in \CS_j}^{}  \big[ \nabla \ln p(x_i \vert \theta)^\top ( \mu_j - x_i)  + \frac{1}{2\gamma}\Vert \mu_j - x_i\Vert^2 \big]  \\ 
= &  \sum_{i \in \CS_j}^{}  \big[ \big( \nabla \ln p(x_i \vert \theta) - \nabla \ln p(\mu_j \vert \theta) +  \nabla \ln p(\mu_j \vert \theta) \big)^\top \\
& ( \mu_j - x_i)  + \frac{1}{2\gamma}\Vert \mu_j - x_i\Vert^2 \big]  \\ 
= &  \sum_{i \in \CS_j}^{}  \big[ \big( \nabla \ln p(x_i \vert \theta) - \nabla \ln p(\mu_j \vert \theta)  \big)^\top ( \mu_j - x_i)\big]  \\
& +   \underbrace{\sum_{i \in \CS_j}^{}  \big[  \nabla \ln p(\mu_j \vert \theta)^\top ( \mu_j - x_i)\big]}_{\text{equal to 0}}  
	%\\&
 + \sum_{i \in \CS_j}^{}  \frac{1}{2\gamma}\Vert \mu_j - x_i\Vert^2\\
= &   \sum_{i \in \CS_j}^{}  \big[ \big( \nabla \ln p(x_i \vert \theta) - \nabla \ln p(\mu_j \vert \theta)  \big)^\top ( \mu_j - x_i)\big]  \\&
+ \sum_{i \in \CS_j}^{}  \frac{1}{2\gamma}\Vert \mu_j - x_i\Vert^2\\
\leq &   \sum_{i \in \CS_j}^{}  \big[ L_1 \Vert x_i - \mu_j \Vert\cdot  L_2 \Vert \mu_j - x_i\Vert \big]  
	%\\&
+ \sum_{i \in \CS_j}^{}  \frac{1}{2\gamma}\Vert \mu_j - x_i\Vert^2\\
= &    \sum_{i \in \CS_j}^{}  (L_1L_2 + \frac{1}{2\gamma})\Vert \mu_j - x_i\Vert^2\\
\leq & n_j (L_1L_2 + \frac{1}{2\gamma}) \max_{i \in \CS_j}\Vert x_i - \mu_j \Vert^2,
\end{aligned}
\end{equation}
where the first inequality uses Assumption~\ref{asmp:smooth} again, the second inequality uses Assumption~\ref{asmp:lipschitz}.

Combining part I and II, we have that 
\begin{equation}
\label{eqn:term_a3}
\begin{aligned}
&  \bigg\vert \sum_{i \in \CS_j}^{}  \big[  \ln p(\mu_j \vert \theta) - \ln p(x_i \vert \theta) \big] \bigg\vert \\
\leq & n_j (L_1L_2 + \frac{1}{2\gamma}) \max_{i \in \CS_j}\Vert x_i - \mu_j \Vert^2,
\end{aligned}
\end{equation}	
Then we sum the identity over $j = 1,\cdots, c$ and have 
\begin{equation}
\label{eqn:part_aa}
\begin{aligned}
& \bigg\vert \ln \frac{p(\theta) \prod_{i=1}^{N} p(x_i \vert\theta)}{p(\theta) \prod_{j=1}^{c} \big( p(\mu_j \vert \theta )\big)^{n_j}} \bigg\vert \\
= & \bigg\vert \sum_{j = 1}^{c} \sum_{i \in \CS_j}^{}  \big[\ln p(x_i \vert \theta) -  \ln p(\mu_j \vert \theta) \big] \bigg\vert\\
\leq  & \sum_{j = 1}^{c} \bigg\vert \sum_{i \in \CS_j}^{}  \big[\ln p(x_i \vert \theta) -  \ln p(\mu_j \vert \theta) \big] \bigg\vert\\
\leq & \sum_{j = 1}^{c} n_j (L_1L_2 + \frac{1}{2\gamma}) \max_{i \in \CS_j}\Vert x_i - \mu_j \Vert^2\\
\leq & (L_1L_2N + \frac{N}{2\gamma}) \delta^2\\
= & K_1 \delta^2, 
\end{aligned}
\end{equation}
where the first inequality follows from the triangle inequality and 
the second inequality directly use the results in Equation~\eqref{eqn:term_a3}.
Until now, we have bounded the first part of $\ln \frac{p(\theta \vert X)  }{\tilde{p}(\theta\vert X)}$ in Equation~\eqref{eqn:log_term}.
For simplicity, we denote $K_1 \triangleq L_1L_2N + \frac{N}{2\gamma}$.

Now we turn our attention to the term ``B'' of last line in Equation~\eqref{eqn:log_term}. 
That is, we attempt to bound the term $\ln \frac{\tilde{Z}}{Z}$.
By taking exponentiation to the both sides of  Equation\eqref{eqn:part_aa} we know that the term $\frac{p(\theta) \prod_{i=1}^{N} p(x_i \vert\theta)}{p(\theta) \prod_{j=1}^{c} \big( p(\mu_j \vert \theta )\big)^{n_j}}$ can be upper- and lower-bounded as
\begin{equation}
\label{eqn:8}
\begin{aligned}
\exp(- K_1 \delta^2) \leq  \frac{p(\theta) \prod_{i=1}^{N} p(x_i \vert\theta)}{p(\theta) \prod_{j=1}^{c} \big( p(\mu_j \vert \theta )\big)^{n_j}}
\leq \exp(K_1 \delta^2). 
\end{aligned}
\end{equation}
	
On the other hand, the term $\ln \frac{\tilde{Z}}{Z}$ can be expanded as follows:
\begin{equation}
\label{eqn:9}
\begin{aligned}
\vert \ln \frac{\tilde{Z}}{Z}\vert = \big\vert \ln \frac{ \int p(\theta) \prod_{i=1}^{N} p(x_i \vert\theta) d\theta}{\int p(\theta) \prod_{j=1}^{c} \big( p(\mu_j \vert \theta )\big)^{n_j}d\theta } \big\vert 
\end{aligned}
\end{equation}
	
Since $p(\theta) \prod_{i=1}^{N} p(x_i \vert\theta) d\theta > 0$ and $p(\theta) \prod_{j=1}^{c} \big( p(\mu_j \vert \theta )\big)^{n_j} > 0$ satisfy for any given $\theta$, it is easy to find that 
\begin{equation}
\label{eqn:10}
\begin{aligned}
& \min_\theta \frac{p(\theta) \prod_{i=1}^{N} p(x_i \vert\theta)}{p(\theta) \prod_{j=1}^{c} \big( p(\mu_j \vert \theta )\big)^{n_j}} \\
\leq & \frac{ \int_\theta p(\theta) \prod_{i=1}^{N} p(x_i \vert\theta) d\theta}{\int_\theta p(\theta) \prod_{j=1}^{c} \big( p(\mu_j \vert \theta )\big)^{n_j}d\theta } = \ln \frac{\tilde{Z}}{Z}\\ 
\leq  & \max_\theta \frac{p(\theta) \prod_{i=1}^{N} p(x_i \vert\theta)}{p(\theta) \prod_{j=1}^{c} \big( p(\mu_j \vert \theta )\big)^{n_j}}.
\end{aligned}
\end{equation}
	
Combining Equation~\eqref{eqn:8} and \eqref{eqn:10}, we have that 
\begin{equation}
\label{eqn:11}
\begin{aligned}
& \exp(- K_1 \delta^2)
\leq  \frac{ \int p(\theta) \prod_{i=1}^{N} p(x_i \vert\theta) d\theta}{\int p(\theta) \prod_{j=1}^{c} \big( p(\mu_j \vert \theta )\big)^{n_j}d\theta }\\ 
& \leq  \exp( K_1 \delta^2).
\end{aligned}
\end{equation}
Taking logorithm to all the terms in above equation, according to the monotonicity of log-function, we have
\begin{equation}
\label{eqn:12}
\begin{aligned}
& - K_1 \delta^2
\leq \ln \frac{ \int p(\theta) \prod_{i=1}^{N} p(x_i \vert\theta) d\theta}{\int p(\theta) \prod_{j=1}^{c} \big( p(\mu_j \vert \theta )\big)^{n_j}d\theta } = \ln \frac{\tilde{Z}}{Z} \\
& \leq K_1 \delta^2.
\end{aligned}
\end{equation}
	
Thus the term $\ln \frac{\tilde{Z}}{Z}$ is bounded.
That is, 
\begin{equation}
\label{eqn:13}
\begin{aligned}
\big\vert \ln \frac{\tilde{Z}}{Z} \big\vert \leq K_1 \delta^2
\end{aligned}
\end{equation}
	
Now both term A and B are bounded in Equation~\eqref{eqn:log_term} are bounded.
Then the term $ \ln \frac{p(\theta \vert X)  }{\tilde{p}(\theta\vert X)}$ can be bounded as
\begin{equation}
\begin{aligned}
& \bigg\vert \ln \frac{p(\theta \vert X)  }{\tilde{p}(\theta\vert X)}\bigg\vert \\
= & \sum_{j=1}^{c} {\bigg[ \sum_{i \in \CS_j}^{}  \big[\ln p(x_i \vert \theta) -  \ln p(\mu_j \vert \theta) \big] \bigg]} + {\ln \frac{\tilde{Z}}{Z}}. \\ 
\leq  &  \sum_{j=1}^{c} {\bigg\vert \sum_{i \in \CS_j}^{}  \big[\ln p(x_i \vert \theta) -  \ln p(\mu_j \vert \theta) \big] \bigg\vert}  + \bigg\vert {\ln \frac{\tilde{Z}}{Z}} \bigg\vert \\ 
\leq & 2K_1 \delta^2.
\end{aligned}
\end{equation}
Based on this, the KL divergence between the exact posterior $p(\theta \vert X)$ and the approximate posterior $\tilde{p}(\theta\vert X)$ is bounded as
\begin{equation}
\begin{aligned}
\text{KL} ( p(\theta \vert X)\Vert \tilde{p}(\theta\vert X) ) = &  \int p(\theta \vert X) \ln \frac{p(\theta \vert X)  }{\tilde{p}(\theta\vert X)} d\theta\\
\leq & \int p(\theta \vert X) \big\vert \ln \frac{p(\theta \vert X)  }{\tilde{p}(\theta\vert X)}\big\vert  d\theta\\
\leq & \int p(\theta \vert X) \max_{\theta}\big\vert \ln \frac{p(\theta \vert X)  }{\tilde{p}(\theta\vert X)}\big\vert  d\theta\\
\leq & \int p(\theta \vert X) 2K_1 \delta^2 d\theta\\
= & 2K_1 \delta^2 = K_0 \delta^2.
\end{aligned}
\end{equation}
where $K_0 = 2K_1 = 2L_1L_2N + \frac{N}{\gamma}$.
Thus, proved.
	
\end{proof}

\section*{Proof of Lemma~\ref{lemma:1}}
\begin{proof}

According to the definition of multivariate normal distribution, we have
\begin{equation*}
\begin{aligned}
& p_1(x\vert\mu_1,\Sigma_1)\\ = & (2\pi)^{-\frac{d}{2}} \vert\Sigma_1 \vert^{-\frac{1}{2}} 
\exp\big( - \frac{1}{2} (x-\mu_1)^\top\Sigma_1^{-1}(x-\mu_1)\big),\\
\end{aligned}
\end{equation*}
\begin{equation*}
\begin{aligned}
& p_2(x\vert\mu_2,\Sigma_2)\\ = & (2\pi)^{-\frac{d}{2}} \vert\Sigma_2 \vert^{-\frac{1}{2}} 
\exp\big( - \frac{1}{2} (x-\mu_2)^\top\Sigma_2^{-1}(x-\mu_2)\big),\\
\end{aligned}
\end{equation*}
and 
\begin{equation*}
\begin{aligned}
& \ln p_1(x\vert\mu_1,\Sigma_1) \\ = & (-\frac{d}{2}) \ln (2\pi) - \frac{1}{2} \ln \vert \Sigma_1 \vert 
- \frac{1}{2} (x-\mu_1)^\top\Sigma_1^{-1}(x-\mu_1).\\
\end{aligned}
\end{equation*}

Then we simplify the KL divergence between $p_1$ and $p_2$ as
\begin{equation*}
\begin{aligned}
& \text{KL}(p_1(x\vert\mu_1,\Sigma_1) \Vert p_2(x\vert\mu_2,\Sigma_2))\\
 = & \int p_1(x\vert\mu_1,\Sigma_1) \ln \frac{p_1(x\vert\mu_1,\Sigma_1)}{p_2(x\vert\mu_2,\Sigma_2)} dx \\
 = & \int p_1 \bigg[- \frac{1}{2} \ln \vert \Sigma_1 \vert 
- \frac{1}{2} (x-\mu_1)^\top\Sigma_1^{-1}(x-\mu_1)\\ 
 & + \frac{1}{2} \ln \vert \Sigma_2 \vert + \frac{1}{2} (x-\mu_2)^\top\Sigma_2^{-1}(x-\mu_2) \bigg]dx\\
 = & - \frac{1}{2} \ln \vert \Sigma_1 \vert + \frac{1}{2} \ln \vert \Sigma_2 \vert \\
& + \frac{1}{2} \int p_1 \bigg[ 
-  (x-\mu_1)^\top\Sigma_1^{-1}(x-\mu_1)\\ 
&  +  (x-\mu_2)^\top\Sigma_2^{-1}(x-\mu_2) \bigg]dx\\
= & - \frac{1}{2} \ln \vert \Sigma_1 \vert + \frac{1}{2} \ln \vert \Sigma_2 \vert  - \frac{1}{2} d \\
& + \frac{1}{2} \big[ \text{tr}[\Sigma_2^{-1}  \Sigma_1] + (\mu_1 - \mu_2)^\top\Sigma_2^{-1} (\mu_1 - \mu_2) \big],\\
\end{aligned}
\end{equation*}

% -1/2*log(det(sigma_1)) + 1/2*log(det(sigma_2)) + 1/2 * (sum(diag(inv(sigma_2)*sigma_1)) + (mu_1-mu_2)' * inv(sigma_2) * (mu_1 - mu_2) ) - 1/2 * fea_dim*class

where the last equality follows from the fact that
\begin{equation*}
\begin{aligned}
& \int p_1(x\vert\mu_1,\Sigma_1) \big[ (x-\mu_2)^\top\Sigma_2^{-1}(x-\mu_2)\big]dx \\
= & \int p_1(x\vert\mu_1,\Sigma_1)
\\&
\big[\text{tr}( \Sigma_2^{-1}xx^\top) - 2 \mu_2 \Sigma_2^{-1} x + \mu_2^\top \Sigma_2^{-1} \mu_2\big] dx\\ 
= & \text{tr}[\Sigma_2^{-1} (\mu_1\mu_1^\top + \Sigma_1)] - 2\mu_2^\top\Sigma_2^{-1}\mu_1 + \mu_2^\top \Sigma_2^{-1} \mu_2 \\
= & \text{tr}[\Sigma_2^{-1}  \Sigma_1] + (\mu_1 - \mu_2)^\top\Sigma_2^{-1} (\mu_1 - \mu_2), \\
\end{aligned}
\end{equation*}

\begin{equation*}
\begin{aligned}
& \int p_1(x\vert\mu_1,\Sigma_1) \big[ (x-\mu_1)^\top\Sigma_1^{-1}(x-\mu_1)\big]dx \\
= & \int p_1(x\vert\mu_1,\Sigma_1) \big[\text{tr}( \Sigma_1^{-1}xx^\top) - 2 \mu_1^\top \Sigma_1^{-1} x\\
&  + \mu_1^\top \Sigma_1^{-1} \mu_1\big] dx\\ 
= & \text{tr}[\Sigma_1^{-1} (\mu_1\mu_1^\top + \Sigma_1)] - 2\mu_1^\top \Sigma_1^{-1}\mu_1 + \mu_1^\top \Sigma_1^{-1} \mu_1 \\
= & d. \\
\end{aligned}
\end{equation*}

\end{proof}

\bibliographystyle{named}
\bibliography{ref}
\end{document}